\documentclass[letterpaper]{article} 
\usepackage{aaai2026}  
\usepackage{times}  
\usepackage{helvet}  
\usepackage{courier}  
\usepackage[hyphens]{url}  
\usepackage{graphicx} 
\usepackage{natbib}  
\usepackage{caption} 
\usepackage{algorithm}
\usepackage{algorithmic}

\usepackage{newfloat}
\usepackage{listings}
\DeclareCaptionStyle{ruled}{labelfont=normalfont,labelsep=colon,strut=off} 
\floatstyle{ruled}
\newfloat{listing}{tb}{lst}{}
\floatname{listing}{Listing}
\title{From Passive Perception to Active Memory: A Weakly Supervised Image Manipulation Localization Framework Driven by Coarse-Grained Annotations}
\author{
   Zhiqing Guo\textsuperscript{1,2}$^{*}$,
   Dongdong Xi\textsuperscript{1}$^{*}$, 
    Songlin Li\textsuperscript{1}$^{\dagger}$,  
    Gaobo Yang\textsuperscript{\rm 3}
}
\affiliations{
   \textsuperscript{\rm 1}School of Computer Science and Technology, Xinjiang University\\
    \textsuperscript{\rm 2}Xinjiang Multimodal Intelligent Processing and Information Security Engineering Technology Research Center \\
    \textsuperscript{\rm 3}College of Computer Science and Electronic Engineering, Hunan University \\
    $^\ast$ These authors contributed equally. \\
   $^\dagger$ Corresponding author: lisl@stu.xju.edu.cn
    
}

\usepackage{bibentry}
\usepackage{amsmath}
\usepackage{amssymb}
\usepackage{booktabs}
\usepackage{makecell}
\usepackage{multirow}

\usepackage{scrextend}
\newcommand\blfootnote[1]{%
  \begingroup
  \renewcommand\thefootnote{}\footnote{#1}%
  \addtocounter{footnote}{-1}%
  \endgroup
}

\begin{document}

\maketitle
\blfootnote{\noindent This is a copy of the copyrighted version accepted at AAAI 2026.}

\begin{abstract}
Image manipulation localization (IML) faces a fundamental trade-off between minimizing annotation cost and achieving fine-grained localization accuracy. 
Existing fully-supervised IML methods depend heavily on dense pixel-level mask annotations, which limits scalability to large datasets or real-world deployment. 
In contrast, the majority of existing weakly-supervised IML approaches are based on image-level labels, which greatly reduce annotation effort but typically lack precise spatial localization.
To address this dilemma, we propose BoxPromptIML, a novel weakly-supervised IML framework that effectively balances annotation cost and localization performance. Specifically, we propose a coarse region annotation strategy, which can generate relatively accurate manipulation masks at lower cost. To improve model efficiency and facilitate deployment, we further design an efficient lightweight student model, which learns to perform fine-grained localization through knowledge distillation from a fixed teacher model based on the Segment Anything Model (SAM). 
Moreover, inspired by the human subconscious memory mechanism, our feature fusion module employs a dual-guidance strategy that actively contextualizes recalled prototypical patterns with real-time observational cues derived from the input. 
Instead of passive feature extraction, this strategy enables a dynamic process of knowledge recollection, where long-term memory is adapted to the specific context of the current image, significantly enhancing localization accuracy and robustness. 
Extensive experiments across both in-distribution and out-of-distribution datasets show that BoxPromptIML outperforms or rivals fully-supervised models, while maintaining strong generalization, low annotation cost, and efficient deployment characteristics. 
\end{abstract}

\begin{links}
    \link{Code}{https://github.com/vpsg-research/BoxPromtIML}
\end{links}

\begin{figure}[t]
\centering
\includegraphics[width=0.9\columnwidth]{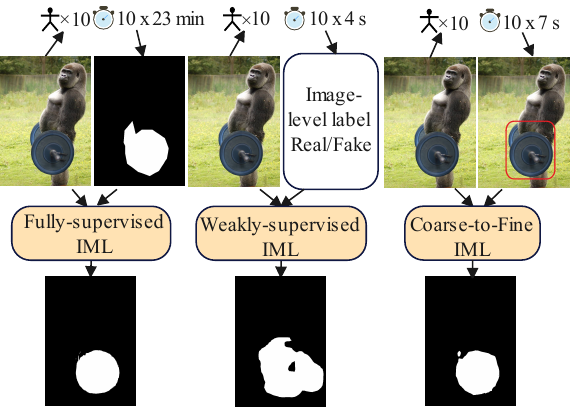} 
\caption{Comparison of annotation cost and supervision quality in different IML paradigms. Pixel-level masks take 23 minutes per image, image-level labels require 4 seconds. Our IML framework from coarse to fine adopts rough boxes which only takes 7 seconds to annotate, while retaining spatial clues.}
\label{fig1}
\end{figure}

\section{Introduction}
Ensuring the authenticity of digital images has become increasingly challenging due to the widespread availability of powerful image editing and generation technologies. These technologies, such as advanced copy-move\cite{copy_move}, cut-paste techniques and GAN-based manipulation, have been misused in misinformation, deepfake content~\cite{deepfake1}, and financial fraud, posing serious risks to social trust and public safety. As a result, IML has emerged as a vital research topic in the fields of  computer vision and multimedia forensics.

Although recent advances have improved IML capabilities, the field still faces a critical trade-off between localization accuracy and annotation cost, as also evidenced by the unified benchmark analysis in ForensicHub~\cite{forensichub}.
A major bottleneck lies in the scarcity of real-world manipulated images with high-quality ground-truth masks~\cite{imd2020}. 
This scarcity is largely due to the prohibitively high cost of manually annotating fine-grained manipulation boundaries, especially when the tampered objects involve intricate textures or complex edges. Creating such pixel-level masks typically requires annotators to perform meticulous per-pixel comparisons along object boundaries, making the process extremely time-consuming and labor-intensive. 
This challenge has significantly hindered the scalability of IML datasets and the broader development of the field. 
To mitigate this issue, some weakly supervised IML methods have been proposed, typically leveraging image-level labels to minimize annotation effort, though they generally lack precise spatial supervision.

To the best of our knowledge, there is currently no quantitative analysis of how long it takes humans to annotate different types of supervision for real-world manipulated images. To address this gap, we conducted a controlled user study involving 10 volunteers, who were asked to annotate 100 tampered images along with their corresponding authentic versions. These image pairs were selected from real-world manipulation examples in the IMD2020~\cite{imd2020} and In-the-Wild~\cite{in-the-wild} datasets. During annotation, participants were allowed to refer to both the tampered and original images to assist in identifying manipulated regions. The labeling tasks included generating pixel-level manipulation masks, image-level real/fake labels, and coarse bounding boxes. The time required for each annotation type was carefully recorded, and cross-validation was performed among participants to ensure annotation quality and consistency. All annotations were conducted using the CVAT platform.

As illustrated in Figure~\ref{fig1}, we compare three representative supervision strategies in IML:
\begin{itemize}
    \item  Fully-supervised IML, which relies on pixel-level manipulation masks and offers the highest localization accuracy, but suffers from prohibitively high annotation cost—up to 23 minutes per image in our user study.
    \item Weakly-supervised IML methods that rely on image-level real/fake labels significantly reduce annotation effort (4 seconds per image), but fails to provide any spatial guidance, often resulting in poor localization.
    \item The proposed Coarse-to-Fine IML, which uses rough bounding box annotations and reduces annotation time to about 7 seconds per image, while still preserving essential spatial information. 
\end{itemize} 

The results above clearly show that rough bounding box annotation offers a compelling cost-performance trade-off, reducing labeling effort by over 98\% compared to full supervision, while preserving essential spatial cues for learning. 
This motivates the need for a middle-ground solution that preserves spatial localization signals without incurring the high cost of pixel-level annotations. 
Thus, we design a coarse-to-fine localization approach in which simple bounding boxes act as inexpensive yet informative prompts to guide the model training.
A frozen SAM~\cite{sam} is then used to transform these coarse prompts into fine-grained pseudo masks. These pseudo masks serve as soft supervision signals to train a lightweight student model, which can independently localize manipulation during inference. 
Inspired by theories of collective memory and selective attention, we design a Memory-Guided Gated Fusion Module (MGFM) to further enhance the localization performance of the student model. 
This module maintains a learnable memory bank that stores prototypical manipulation patterns, and employs a gating mechanism to selectively fuse multi-scale features based on their relevance. 
Memory bank is structurally advantageous as it decouples knowledge aggregation from the network's weights. By storing and averaging feature activations from past iterations, it creates a more stable and diverse global prior of manipulation archetypes than what can be implicitly learned by standard attention mechanisms. This acts as a strong regularizer that forces the model to reconcile real-time evidence with this robust prior, which is critical for preventing overfitting and improving OOD generalization. 

Our main contributions are summarized as follows:
\begin{itemize}
    \item We propose a novel coarse-to-fine weak supervision paradigm for IML, where only low-cost coarse bounding boxes are required instead of expensive pixel-level masks. A frozen SAM model transforms these prompts into high-quality pseudo masks, which are used to distill knowledge into a compact student model.
    \item We design a memory-guided gated fusion module that retrieves and integrates prototypical manipulation pattern, enhancing the student model’s ability to localize subtle and diverse manipulation.
    \item Extensive experiments on both in-distribution and out-of-distribution datasets demonstrate that our method not only comparable to or superior to several fully-supervised state-of-the-art models, but also exhibits strong generalization and fast convergence under limited supervision,  which provides a new paradigm for the design of weak supervision frameworks.
\end{itemize}

\begin{figure*}[t]
\centering
\includegraphics[width=1.0\textwidth]{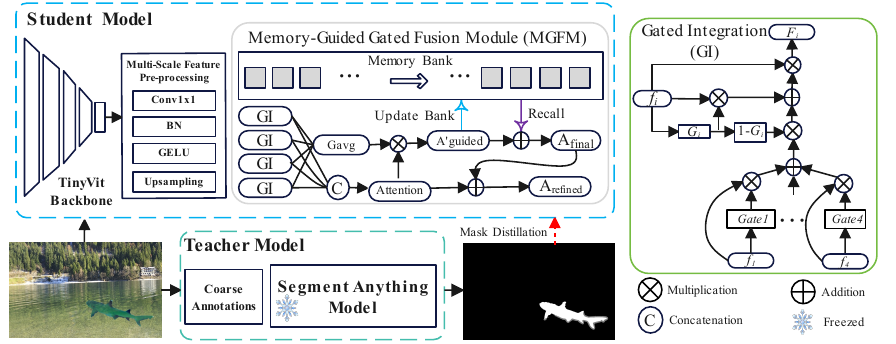} 
\caption{Overview of the proposed framework for fine-grained manipulation localization using coarse prompts. A frozen SAM generates pseudo-masks from coarse annotations (e.g., bounding boxes) as soft supervision. The student model learns to replicate these masks via knowledge distillation. To improve localization, we design a Memory-Guided Gated Fusion Module (MGFM) that fuses multi-scale features with guidance from both real-time and memory-recalled signals. This enables refined mask prediction using only coarse supervision.
 }
\label{fig2}
\end{figure*}

\section{Related Work}

\subsection{Fully-Supervised Image Manipulation Localization}
As image manipulations grew more sophisticated, fully supervised IML approaches empowered by large-scale pixel-level annotations showed remarkable capabilities in localizing tampered regions.
TruFor~\cite{trufor} leveraged contrastive learning to extract noise fingerprints and refined anomaly localization by correcting confidence maps, thereby reducing false positives. 
PIM~\cite{pim} enhanced generalization by modeling pixel-level inconsistencies caused by disruptions to natural image signal processing pipelines. 
Mesorch~\cite{mesorch} addressed the limitation of relying solely on low-level visual traces by integrating both micro- and macro-level features into a unified mesoscopic representation, offering a more holistic approach to manipulation detection.

\subsection{Weakly-Supervised Image Manipulation Localization}
To reduce reliance on costly pixel-level annotations, the WSCL framework~\cite{wscl} introduced a novel paradigm for training IML models using only image-level binary labels. By enforcing consistency across multiple sources and spatial patches, WSCL achieved strong image-level detection performance in both in-distribution and out-of-distribution settings. 
Another representative weakly supervised IML method was the SCAF~\cite{li2025beyond}, which employed scribble annotations to enhance supervision and achieved strong performance, even surpassing fully-supervised methods on both in-domain and out-of-domain datasets.

Fully-supervised IML methods offer accurate localization but rely heavily on dense annotations, limiting scalability. 
Weakly-supervised methods that rely solely on image-level labels reduce annotation cost but generally struggle to achieve precise localization.
The SCAF sets a new performance benchmark for IML by outperforming fully supervised counterparts through the use of scribble-based annotations. 
However, its relatively high annotation cost (around 20 seconds per image), slower convergence, and large model size and computational overhead limit its practicality for resource-constrained scenarios.
To address these limitations, we propose a lightweight, efficient, and fast-converging weakly-supervised IML framework that further reduces annotation effort, offering a practical solution for real-world image forensics.

\section{Method}
This section presents our proposed weakly-supervised framework for  IML. The core of our approach is built upon a teacher–student knowledge distillation paradigm, aiming to train an efficient and accurate localization model without requiring any pixel-level ground-truth masks. An overview of the framework is illustrated in Figure~\ref{fig2}.

Our key idea is to leverage a large-scale foundation segmentation model as the teacher, which possesses powerful zero-shot segmentation capabilities. Given a coarse prompt (e.g., a rough bounding box indicating the manipulated region), the teacher model can generate high-quality fine-grained masks. These masks, although not manually annotated, serve as pseudo ground truth for supervising a lightweight student model. The student is trained to predict tampered regions directly from the input image, without relying on any external prompt during inference. 
\subsection{Teacher Model: Prompt-Based High-Fidelity Mask Generation}
To avoid reliance on pixel-level annotations, we employ the pre-trained SAM as a frozen teacher model. Given an input image \( I \in \mathbb{R}^{3 \times H \times W} \) and a coarse bounding box \( B \in \mathbb{R}^4 \) indicating the manipulated region, SAM generates a high-quality binary mask \( \hat M_{teacher} \in \{0,1\}^{H \times W} \) via its prompt encoder and mask decoder.
These automatically produced pseudo masks provide fine-grained spatial supervision without requiring manual labeling. By leveraging SAM’s zero-shot segmentation capability, our framework transforms low-cost box-level prompts into detailed masks, enabling effective supervision while minimizing annotation cost.
The student model is trained using these pseudo masks, as described in the following section.

\subsection{Memory-Guided Gated Fusion}
To enable our student model to learn precise localization from the teacher's pseudo-masks, a powerful and robust feature fusion mechanism is required. 
Standard feature fusion networks often fall short because they treat all features passively and lack a mechanism to handle the diversity of manipulations, from subtle splicing artifacts to novel AI generated content.
Our key insight draws inspiration from the psychological concept of the collective unconscious, which posits that human understanding is shaped not only by personal experience but also by a shared, inherited reservoir of archetypal patterns.
Translating this to our domain, we argue that a robust manipulate localization model should not rely solely on analyzing a single image in isolation. Instead, it must emulate this dual cognitive process. It needs to combine the the meticulous analysis of immediate evidence within the current image and the recall of prototypical manipulation patterns which learned collectively from a vast corpus of examples. 
To realize this, we designed the Memory-Guided Gated Fusion Module  (MGFM), as illustrated in Figure~\ref{fig2}. 
The MGFM operationalizes this dual-pronged approach. 
Its Gated Integration mechanism dynamically assesses multi-scale features to identify immediate contextual anomalies, while its Memory Bank recalls prototypical tampering patterns to provide historical context. 
This entire cognitive process is predicated on the module's ability to effectively interpret the initial evidence, which is provided by the backbone network in the form of multi-scale features.

\paragraph{Multi-Scale Feature Pre-Processing.}
We adopt Tiny-ViT~\cite{tinyvit} as the backbone to extract four multi-scale features ${F_1, F_2, F_3, F_4}$. To enable effective fusion, we unify their dimensions and resolutions to obtain aligned features ${F'_1, F'_2, F'_3, F'_4}$.

\paragraph{Gated Integration and Memory-Guided Refinement.}
The workflow within the MGFM begins at the Gated Integration (GI) stage. This stage is responsible for intelligently merging the multi-scale features $\{F'_1, ..., F'_4\}$. For each feature map $F'_i$, a corresponding gate map $G_i$ is generated to learn its spatial importance:
\begin{align}
    G_i = \sigma(\text{Conv}_{1 \times 1}(F'_i))
\end{align}

These gates then steer a dynamic fusion strategy. For each feature $F'_i$, a preliminary fused feature $F''_{i}$ is computed, which combines the feature with a weighted aggregation of all other features:
\begin{align}
    F''_{i} = (G_i \odot F'_i) + ((1 - G_i) \odot \sum_{j \neq i} (G_j \odot F'_j))
\end{align}
where $\odot$ denotes the element-wise Hadamard product. This result is further modulated to enhance the original feature's contribution:
\begin{align}
    \hat{F}_i = F'_i \odot F''_{i}
\end{align}
Finally, all modulated features $\{\hat{F}_1, \hat{F}_2, \hat{F}_3, \hat{F}_4\}$ are concatenated and passed through a convolutional block to produce a single, richly integrated feature map, $F_{fused}$.

Simultaneously, this stage yields an average gate map, $G_{avg} = \frac{1}{4}\sum_{i=1}^{4} G_i$. With $F_{fused}$ and $G_{avg}$ available, the module proceeds to its core memory-guided refinement stage. This stage is driven by a dual-guidance mechanism. First, an efficient attention mechanism computes a base attention map, $A'_{base}$, from $F_{fused}$. The refinement of this attention is then guided by two synergistic sources of information:
\begin{itemize}
    \item Real-time Gate Prior: The average gate map, $G_{avg}$, provides a direct spatial saliency prior derived from the current input.
    \item Long-term Memory Prior: Prototypical tampering patterns are recalled from a Memory Bank, which provides a historical knowledge prior, $\bar{A}_{mem}$.
\end{itemize}
These two priors work in concert to produce the final, memory-informed attention map, $A_{final}$, which embodies both immediate analysis and long-term experience:
\begin{align}
    A_{final} = \alpha (A'_{base} \odot G_{avg}) + (1-\alpha) \bar{A}_{mem}
\end{align}
where $\alpha$ is a balancing factor. This final attention map $A_{final}$ is then used to re-weight the feature representations, generating a refined output, $A_{refined}$, through a residual connection, which ultimately forms the model's final prediction.

\subsection{Training Objective}
Our training objective is designed to enable the student model to accurately localize manipulated regions without any human-annotated pixel-level masks. To achieve this, we adopt a pseudo-supervised distillation strategy, where the high-fidelity masks generated by the teacher model serve as soft supervision signals. The student model, which takes only the raw image \( I \) as input, predicts a manipulation localization map \( A_{refined} \in \left[ 0,1\right] ^ {H \times W} \). This prediction is supervised by the pseudo mask from the teacher using a standard binary cross-entropy loss:
\begin{align}
    \mathcal{L}_{loss} = \text{BCE}(A_{\text{ refined}}, \hat{M}_{\text{teacher}})
\end{align}
where BCE is computed element-wise across all pixels. This loss encourages the student model to replicate the teacher’s mask quality, even though it receives no prompt or bounding box during inference.

\begin{table*}[ht]
    \centering
    \small
    \setlength{\tabcolsep}{1mm}
    \begin{tabular}{@{}l c c ccccc c cccccc@{}}
        \toprule
        \multirow{2}{*}{\textbf{Method}} &       
        \multirow{2}{*}{\textbf{Sup.}} &  
        \multirow{2}{*}{\textbf{Epoch}} & \multicolumn{5}{c}{\textbf{In-distribution (IND) F1}} & & 
        \multicolumn{6}{c}{\textbf{Out-of-distribution (OOD) F1}} \\
        \cmidrule(lr){4-8} \cmidrule(lr){10-15}
        & & &  NIST16 & CASIAv1 & Columbia & Coverage & \textbf{Avg.} & & CocoGlide & In-the-Wild & Korus & DSO & IMD2020 & \textbf{Avg.} \\
        \midrule

        \multirow{2}{*}{PSCC-Net} & \multirow{3}{*}{Full}
        & 10 & 0.379 & 0.444 & 0.783 & 0.407 & 0.503 & & \pmb{0.448} & \pmb{0.454} & 0.218 & \pmb{0.349} & 0.331 & \pmb{0.360} \\
        & & 20 & 0.442 & 0.434 & 0.855 & 0.407 & 0.535 & & 0.362 & \pmb{0.425} & 0.221 & \pmb{0.329} & 0.332 & 0.334 \\
        (TCSVT'22) & & 70 & 0.509 & 0.503 & 0.905 & 0.424 & 0.585 & & 0.344 & 0.407 & 0.229 & 0.284 & 0.350 & 0.323 \\

        \midrule
        \multirow{2}{*}{Trufor} & \multirow{3}{*}{Full}
        & 10 & 0.353 & 0.394 & 0.756 & 0.409 & 0.478 & & 0.373 & 0.390 & 0.236 & 0.216 & 0.298 & 0.303 \\
        & & 20 & 0.407 & 0.477 & 0.866 & 0.402 & 0.538 & & 0.261 & 0.287 & 0.182 & 0.102 & 0.348 & 0.236 \\
        (CVPR'23) & & 70 & 0.584 & 0.603 & 0.953 & 0.435 & 0.644 & & 0.287 & 0.310 & 0.200 & 0.165 & 0.375 & 0.267 \\

        \midrule
        \multirow{2}{*}{MFI-Net} & \multirow{3}{*}{Full}
        & 10 & 0.538 & 0.433 & 0.925 & 0.229 & 0.531 & & 0.314 & 0.274 & 0.188 & 0.193 & 0.300 & 0.254 \\
        & & 20 & \pmb{0.629} & 0.409 & 0.944 & 0.409 & 0.598 & & 0.245 & 0.310 & 0.189 & 0.254 & 0.330 & 0.266 \\
        (TCSVT'24) & & 70 & 0.817 & 0.524 & 0.938 & 0.497 & 0.694 & & 0.283 & 0.300 & 0.186 & 0.194 & 0.329 & 0.258 \\

        \midrule
        \multirow{2}{*}{Mesorch} & \multirow{3}{*}{Full}
        & 10 & 0.335 & 0.434 & 0.871 & 0.362 & 0.501 & & 0.294 & 0.278 & 0.168 & 0.201 & 0.201 & 0.228 \\
        & & 20 & 0.427 & 0.485 & 0.856 & 0.408 & 0.544 & & 0.279 & 0.242 & 0.160 & 0.162 & 0.250 & 0.219 \\
        (AAAI'25) &  & 70 & 0.802 & 0.703 & 0.981 & 0.531 & 0.754 & & 0.218 & 0.299 & 0.182 & 0.160 & 0.346 & 0.241 \\

        \midrule
        \multirow{2}{*}{SparseVit} & \multirow{3}{*}{Full}
        & 10 & 0.402 & 0.486 & 0.889 & 0.286 & 0.516 & & 0.232 & 0.288 & 0.176 & 0.247 & 0.310 & 0.250 \\
        & & 20 & 0.417 & 0.506 & 0.916 & 0.410 & 0.562 & & 0.262 & 0.285 & 0.182 & 0.214 & 0.330 & 0.255 \\
        (AAAI'25) & & 70 & 0.616 & 0.557 & 0.958 & 0.546 & 0.669 & & 0.311 & 0.333 & 0.193 & 0.268 & 0.381 & 0.297 \\

        \midrule
        \multirow{1}{*}{PIM} & \multirow{2}{*}{Full}
        & 10 & 0.460 & \pmb{0.570} & \pmb{0.938} & \pmb{0.493} & \pmb{0.615} & & 0.369 & 0.410 & 0.198 & 0.273 & \pmb{0.400} & 0.330 \\
        (TPAMI'25) & & 20 & 0.587 & 0.548 & \pmb{0.954} & \pmb{0.504} & \pmb{0.648} & & \pmb{0.481} & 0.392 & 0.203 & 0.316 & \pmb{0.395} & \pmb{0.357} \\

        \midrule
        \multirow{2}{*}{\textbf{Ours}} & \multirow{2}{*}{\textbf{Weak}}
        & 10 & \pmb{0.566} & 0.538 & 0.883 & 0.376 & 0.591 & & 0.249 & 0.311 & \pmb{0.237} & 0.169 & 0.301 & 0.253 \\
        & & 20 & 0.618 & \pmb{0.552} & 0.903 & 0.403 & 0.619 & & 0.289 & 0.353 & \pmb{0.249} & 0.205 & 0.328 & 0.285 \\

        \bottomrule

    \end{tabular}%
    \caption{Performance (F1 score with fixed threshold: 0.5) comparison across different training epochs for both IND and OOD datasets. Sup.: Full (pixel-level masks), Weak (bounding boxes). Our method shows strong performance on multiple benchmarks using only 20 epochs. The results also serve as a baseline to analyze overfitting and convergence speed across models.}
    \label{tab:combined_performance}
\end{table*}

\section{Experiments}
\subsection{Implementation Details}
Detailed training configurations and implementation details are provided in Appendix A.I.
Training is performed on a composite dataset constructed from four widely used IML datasets: CASIAv2~\cite{casia}, Coverage~\cite{coverage}, Columbia~\cite{columbia}, and NIST16~\cite{nist16}. All training samples are manipulated images with associated coarse bounding boxes used as prompts for the teacher model. 
Our model is trained end-to-end in a weakly supervised manner, without using any real pixel-level ground-truth masks. The distribution of manipulation types and the specific train/test splits are summarized in Appendix A.II.
For a fair comparison, all baseline models are retrained on the same training set with identical data distributions. Their implementations are adapted from official GitHub repositories, the IMDLBenco~\cite{imdlbenco}, or reimplemented according to the original papers. This ensures that performance differences reflect true model capability rather than inconsistencies in training data.

During evaluation, only the student model is used for inference. No test-time prompts or annotations are provided; the model directly predicts tampered regions from input images. We evaluate on two categories of datasets:
\begin{itemize}
    \item In-distribution (IND): Datasets similar in manipulation types and image styles to the training data.
    \item Out-of-distribution (OOD): Unseen datasets with different manipulation styles or domains, including CocoGlide~\cite{cocoglide}, DSO~\cite{dso}, IMD2020, Korus~\cite{korus}, and In-the-Wild.
\end{itemize}

\subsection{Evaluation Metrics}
While pixel-level F1 and AUC are widely used metrics in IML, recent research indicates that pixel-level AUC often overestimates confidence in these tasks~\cite{imdlbenco}. Consequently, we have opted to evaluate all our experiments using the F1-score with a fixed threshold of 0.5. To comprehensively evaluate model efficiency, we also report the number of parameters and FLOPs for each method.

\subsection{Comparison with Fully-Supervised Methods}
To fairly assess the generalization dynamics of different models, we evaluate their performance at 10, 20, and 70 training epochs. This setting enables a direct comparison of how quickly each model learns and how well it generalizes, especially to OOD scenarios.
All methods are trained on our unified mixed training set for consistent comparison.
Results are summarized in Table~\ref{tab:combined_performance}.
Among all evaluated models, some full-supervised methods like Mesorch and MFI-Net~\cite{mfi-net} show strong IND performance when given sufficient training. For example, Mesorch achieves an average F1 of 0.754 at 70 epochs, and MFI-Net reaches 0.694. These results demonstrate the effectiveness of their sophisticated architectures in modeling fine-grained manipulation traces. 
Despite not having access to ground-truth masks, our method achieves a competitive F1 score of 0.619 at only 20 epochs. This surpasses several strong baselines including Trufor, PSCC-Net~\cite{pscc-net}, and SparseViT~\cite{sparsevit}, demonstrating the effectiveness of our pseudo-supervision strategy.

Generalization to unseen domains remains a significant challenge for IML models. While several full-supervised methods perform well in the IND setting, their robustness in OOD scenarios is often limited. 
For example, TruFor and PSCC-Net demonstrate a concerning trend: their average F1 scores on OOD datasets decrease as training epochs increase. Specifically, TruFor drops from 0.303 (10 epochs) to 0.267 (70 epochs), and PSCC-Net degrades from 0.360 to 0.323 over the same range. This indicates that these models are prone to overfitting and may struggle to generalize beyond the training distribution.
Similarly, Mesorch, though highly effective on IND datasets, shows limited OOD performance even after extended training, improving only marginally from 0.228 to 0.241. 
In contrast, methods like PIM and SparseViT exhibit more stable improvements across training epochs. PIM steadily increases its average OOD F1 from 0.330 to 0.357 between 10 and 20 epochs, while SparseViT grows from 0.250 to 0.297 over 70 epochs, reflecting better generalization potential.
Our method also demonstrates consistent gains in OOD performance, increasing from 0.253 to 0.285 between 10 and 20 epochs, despite being trained without real masks. 

Overall, these results highlight the trade-offs between different design choices in IML models. 
Fully-supervised methods such as Mesorch, TruFor and PSCC-Net exhibit high performance on IND datasets but suffer from poor generalization to unseen domains. This degradation with more training epochs suggests overfitting to dataset-specific patterns, possibly due to their reliance on finely annotated pixel-level masks, which can induce semantic bias during training. 
On the other hand, models like PIM and SparseViT demonstrate relatively better OOD consistency, indicating that stronger architectural priors or training regularizations can improve cross-domain robustness. 
However, their reliance on full supervision and complex backbones still limits scalability.
In contrast, our method achieves a favorable balance across multiple dimensions: it performs competitively on IND benchmarks, maintains stable improvements on OOD datasets, and requires no ground-truth masks. The consistent upward trend in OOD performance, even under limited supervision, highlights the generalization capability of our coarse-to-fine pseudo-labeling framework.

\begin{table}[htb]
\centering
\small
\setlength{\tabcolsep}{1mm}
\begin{tabular}{@{}l c ccccc@{}} 
\toprule
\multirow{2}{*}{\textbf{Method}} & \multirow{2}{*}{\textbf{Pub.}} & \multicolumn{5}{c}{\textbf{In-Distribution}}\\ \cline{3-7} 
& & {{NC16}} &{{C1}} & {{Col}} & {{Cov}} & {{Avg.}} \\ \toprule
WSCL & ICCV'23 & 0.111 & 0.140 & 0.524 & 0.180 & 0.239 \\
EdgeCAM & ESWA'24 & 0.254 & 0.301 & 0.470 & 0.262 & 0.322 \\
SOWCL & ICASSP'25 & 0.288 & 0.334 & 0.385 & 0.239 & 0.312 \\
WSCCL	 &KBS'25   & 0.278 & 0.349 & 0.516 & 0.281 & 0.356 \\
SCAF & Arxiv'25 & 0.226 & 0.530 & 0.442 & 0.400 & 0.400 \\
Ours & - & \pmb{0.618} & \pmb{0.552} & \pmb{0.903} & \pmb{0.403} & \pmb{0.619} \\
\bottomrule
\end{tabular}%
\caption{Comparison with other weakly supervised methods.
NIST16, CASIAv1, Columbia and Coverage are abbreviated as NC16, C1, Col and Cov, respectively.}
\label{weakly_supervised_method}
\end{table}

\subsection{Comparison with Weakly-Supervised Methods}
Table~\ref{weakly_supervised_method} presents a comparison between our method and existing weakly supervised IML approaches on the IND benchmark. 
Due to code availability, only WSCL and SCAF are retrained on our benchmark using the same 20 epoch training schedule. 
Results for other methods are taken directly from their respective papers. 
While direct comparisons may be influenced by differences in training settings, they still offer a meaningful indication of relative performance under weak supervision.

Our method achieves a substantial performance gain over most existing weakly supervised methods. It reaches an average F1 score of 0.619, markedly outperforming WSCL (0.239), EdgeCAM~\cite{EdgeCAM} (0.3218), SOWCL~\cite{SOWCL} (0.312), and WSCCL~\cite{WSCCL} (0.356). These results validate the effectiveness of our coarse-to-fine pseudo-mask supervision strategy, which utilizes spatial prompts to guide learning while keeping annotation cost minimal.
Notably, the recently proposed SCAF method which based on scribble annotations has shown strong localization performance and is capable of surpassing fully supervised methods when trained to full convergence. 
However, SCAF relies on a more intensive form of supervision, with higher annotation time and introduces a larger model size and computational cost. 
In our controlled 20-epoch setting, SCAF does not reach its optimal performance, highlighting a trade-off between supervision strength and training efficiency.
In summary, while scribble-based supervision enhances localization quality, our method offers a more practical alternative that combines low annotation cost, fast convergence, and strong localization performance under minimal supervision.

\begin{table}[htbp]
\centering
\small
\setlength{\tabcolsep}{1mm}
\begin{tabular}{cccc}
\toprule
Method  &Size    & Params(M) & Flops(G) \\
\midrule
Mesorch   & \( 512\times512\)  & 85.8     & 124.9   \\
PIM       & \( 512\times512\)  & 152.5    & 682.9   \\
Trufor    & \( 512\times512\)  & 68.70     & 236.5   \\
PSCC-Net  & \( 256\times256\)  & \pmb{3.7}      & 45.7    \\
SparseVit & \( 512\times512\) & 50.3     & 46.2    \\
MFI-Net        & \( 512\times512\) & 32.54     & 36.25    \\
SCAF      &  \( 512\times512\)  & 27.57    &  35.39 \\
Ours       & \( 224\times224\) & 5.5      & \pmb{1.4}     \\
\bottomrule
\end{tabular}
\caption{Comparison of model size and computational cost.}
\label{tab:comparison_flops}
\end{table}

\begin{table}
\centering
\small
\setlength{\tabcolsep}{1mm}
\begin{tabular}{c|ccccc}
\hline \multirow{2}{*}{ Method } & \multicolumn{5}{c}{ \textbf{App} } \\
\cline { 2 - 6 } & None & Facebook & WeiBo & WeChat & WhatsApp \\
\hline 
WSCL    & 0.349 & 0.124 & 0.133 & 0.066 & 0.123 \\
MFI-Net & 0.524 & 0.449 & 0.455 & 0.363 & 0.474 \\
SparesViT & 0.557 & 0.493 &  0.529 & 0.365 & 0.506 \\
PIM   & 0.548 &0.581 &0.566 &0.505 &0.585\\
Mesorch& \pmb{0.703}  &  \pmb{0.671}  & \pmb{0.655}  &   \pmb{0.583}  &  \pmb{0.677}  \\
Ours &  0.552 & 0.532 & 0.536  &  0.477  &  0.530   \\
\hline
\end{tabular}
\caption{Robustness experiments on online social networks.}
\label{tab:roust}
\end{table}

\subsection{Model Efficiency: Parameters and FLOPs}
We evaluate model efficiency in terms of parameter count and FLOPs. As shown in Table~\ref{tab:comparison_flops}, our model achieves the lowest FLOPs (1.4G) and the second-smallest number of parameters (5.5M) among all methods. 
Despite its lightweight design and lower resolution input, it maintains competitive localization performance. 

\begin{figure}[t]
\centering
\includegraphics[width=0.47\textwidth]{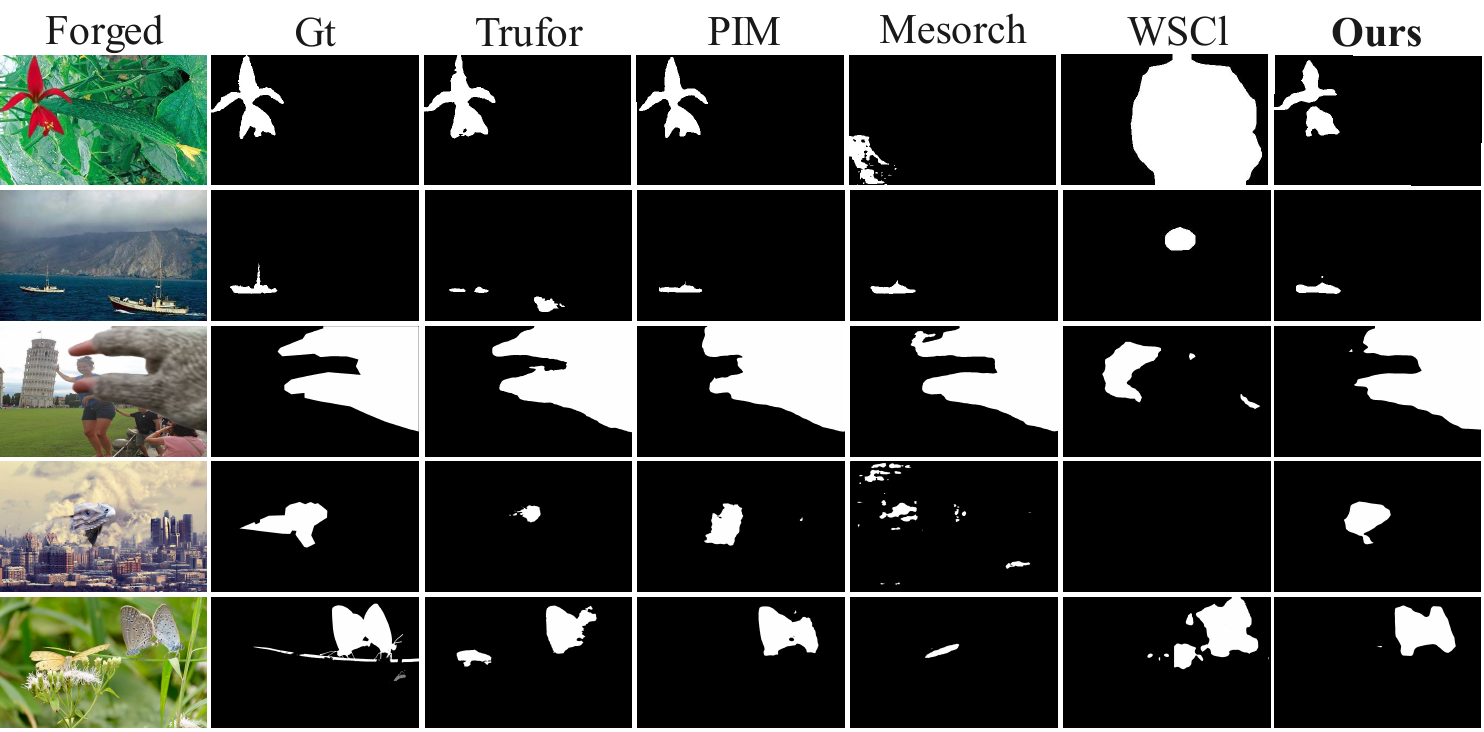} 
\caption{Qualitative comparison of manipulation localization results on both IND (top two rows) and OOD (bottom three rows) examples.}
\label{fig3}
\end{figure}

\begin{table*}[t]
\centering
\renewcommand{\arraystretch}{0.70} 
\small
\setlength{\tabcolsep}{1mm}
\renewcommand{\arraystretch}{1.1}
\begin{tabular}{l|ccccc|cccccc}
\toprule
\multirow{2}{*}{\textbf{Model Configuration}} & \multicolumn{5}{c|}{\textbf{In-Distribution (IND) F1}} & \multicolumn{6}{c}{\textbf{Out-of-Distribution (OOD) F1}} \\
& NIST16 & CASIAv1 & Columbia & Coverage & \textbf{Avg.} & CocoGlide & In-the-Wild & Korus & DSO & IMD2020 & \textbf{Avg.} \\
\midrule
Baseline & 0.599 & 0.532 & 0.899 & 0.367 & 0.599 & 0.233 & 0.339 & 0.223 & 0.148 & 0.318 & 0.252 \\
w/o Memory & 0.594 & 0.479 & 0.891 & 0.340 & 0.576 & 0.267 & 0.185 & 0.241 & 0.145 & 0.298 & 0.227 \\
w/o Gate Prior ($G_{avg}$) & 0.573 & 0.519 & 0.919 & 0.463 & 0.618 & 0.259 & 0.333 & 0.254 & 0.218 & 0.328 & 0.277 \\
w/o Gating & 0.571 & 0.495 & 0.875 & 0.374 & 0.579 & 0.216 & 0.312 & 0.240 & 0.172 & 0.308 & 0.250 \\
Ours (Full Model) & 0.618 & 0.552 & 0.903 & 0.403 & \textbf{0.619} & 0.289 & 0.353 & 0.249 & 0.205 & 0.328 & \textbf{0.285} \\
\bottomrule
\end{tabular}%
\caption{Ablation study of our proposed MGFM on in-distribution (IND) and out-of-distribution (OOD) datasets. We report the F1 score for each dataset and the average F1 scores for IND and OOD sets. The best results are highlighted in \textbf{bold}.}
\label{tab:ablation_study}
\end{table*}

\subsection{Robustness Against Social Media Compression}
To assess real-world robustness, we evaluate models on tampered images recompressed by common social platforms (Facebook, Weibo, WeChat, WhatsApp). 
As shown in Table~\ref{tab:roust}, the weakly-supervised baseline WSCL suffers severe performance drops, with F1 scores falling below 0.15 in all compressed settings. Fully-supervised model Mesorch show the strongest overall performance under compression. Our method maintains strong robustness across platforms, outperforming several fully-supervised models such as MFI-Net and SparseViT, despite relying only on coarse annotations. These results highlight the strong generalization and practicality of our approach under real-world degradation.

\subsection{Qualitative Comparisons}
Figure~\ref{fig3} presents a qualitative comparison of our method against several baselines on both IND and OOD samples, highlighting its effectiveness and generalization capability.

Performance on IND Data: On IND samples (Row 1), our method generates masks that are highly consistent with the ground truth in terms of accuracy and compactness. Its performance is visually indistinguishable from top fully-supervised methods like Trufor and PIM, while clearly surpassing the broad and inaccurate masks produced by the weakly-supervised baseline (WSCL).

Robustness on OOD Data: The key advantage of our method is its robustness on challenging OOD samples (Rows 3-5). For example, in the butterfly case (Row 5), our result is the most complete. Even in a highly ambiguous scene (Row 4), our method captures the manipulated region more clearly than the alternatives. In contrast, while fully-supervised methods perform well on some OOD instances (e.g., Trufor in Row 3), they often struggle with incomplete detection or weak activations in others (Rows 4, 5). 
\subsection{Ablation Study}

We conduct ablation studies to validate the effectiveness of each key component within our proposed MGFM, with results presented in Table \ref{tab:ablation_study}. Our Baseline model consists of the backbone network followed by a simple FPN-style decoder. 
As shown in the table, our full model significantly outperforms the Baseline, confirming the overall superiority of the MGFM architecture. We analyze the contribution of each core component below.
\begin{itemize}
    \item Memory Bank: Removing the memory component (`w/o Memory`) causes the most severe performance degradation, especially on the OOD sets (F1 score drops from 0.285 to 0.227). Similarly, the IND score decreased from 0.619 to 0.576. This strongly proves that recalling prior knowledge of tampering patterns from memory is crucial for the model's performance and generalization ability, especially when confronted with diverse and unseen data from out-of-distribution domains.
    \item Gated Integration: Replacing our GI mechanism with simple feature concatenation (`w/o Gating`) also leads to a substantial performance drop, with the average F1 scores decreasing to 0.579 (IND) and 0.250 (OOD). This indicates that our sophisticated gating logic, which dynamically weighs and modulates features from different scales, is superior to simple concatenation. It generates more discriminative fused features, laying a more solid foundation for the subsequent refinement stages.
    \item Gate Prior ($G_{avg}$): The impact of the gate prior (`w/o Gate Prior`) is most evident on OOD datasets. While its effect on IND sets is marginal (0.619 to 0.618), its necessity for robust generalization to unseen data is clear (0.285 to 0.277). We hypothesize that for IND data, which is similar to the training distribution, the model can heavily rely on the powerful long-term patterns recalled from the memory bank. However, when facing more challenging and unfamiliar OOD data, the real-time spatial prior generated by $G_{avg}$ becomes more critical. It provides immediate, input-specific guidance that helps the model adapt to novel scenes. 
\end{itemize}
In conclusion, all ablated components are proven to be effective and essential. They synergistically form our proposed dual-guidance mechanism (real-time gate prior + long-term memory prior), which is key to achieving robust and accurate manipulation localization.

\section{Conclusion}
In this paper, we propose a novel weakly supervised framework for IML that eliminates the need for pixel-level mask annotations. By leveraging a powerful segmentation foundation model (SAM) with coarse prompts, we generate fine-grained pseudo masks at minimal annotation cost, which in turn guide the training of a lightweight student model. To further boost localization performance, we propose a memory-guided gated fusion module that stores prototypical tampering patterns, improving the model’s generalization and localization accuracy for complex manipulations. Extensive experiments on both in-distribution and out-of-distribution datasets demonstrate that our method achieves competitive or superior localization performance compared to fully supervised methods, while offering strong generalization, faster convergence, and deployment efficiency. In summary, our approach provides a practical and scalable solution for IML with minimal supervision.

\appendix
\section{Acknowledgments}
This work was supported in part by the National Natural Science Foundation of China under Grant 62302427, Grant 62462060, and Grant 62472368, in part by the Natural Science Foundation of Xinjiang Uygur Autonomous Region under Grant 2023D01C175.

\bibliography{aaai2026}

@inproceedings{mesorch,
  title={Mesoscopic insights: orchestrating multi-scale \& hybrid architecture for image manipulation localization},
  author={Zhu, Xuekang and Ma, Xiaochen and Su, Lei and Jiang, Zhuohang and Du, Bo and Wang, Xiwen and Lei, Zeyu and Feng, Wentao and Pun, Chi-Man and Zhou, Ji-Zhe},
  booktitle={Proceedings of the AAAI Conference on Artificial Intelligence},
  volume={39},
  number={10},
  pages={11022--11030},
  year={2025}
}

@article{pim,
  title={Pixel-inconsistency modeling for image manipulation localization},
  author={Kong, Chenqi and Luo, Anwei and Wang, Shiqi and Li, Haoliang and Rocha, Anderson and Kot, Alex C},
  journal={IEEE Transactions on Pattern Analysis and Machine Intelligence},
  year={2025},
  publisher={IEEE}
}

@inproceedings{sparsevit,
  title={Can we get rid of handcrafted feature extractors? sparsevit: Nonsemantics-centered, parameter-efficient image manipulation localization through spare-coding transformer},
  author={Su, Lei and Ma, Xiaochen and Zhu, Xuekang and Niu, Chaoqun and Lei, Zeyu and Zhou, Ji-Zhe},
  booktitle={Proceedings of the AAAI Conference on Artificial Intelligence},
  volume={39},
  number={7},
  pages={7024--7032},
  year={2025}
}

@article{pscc-net,
  title={PSCC-Net: Progressive spatio-channel correlation network for image manipulation detection and localization},
  author={Liu, Xiaohong and Liu, Yaojie and Chen, Jun and Liu, Xiaoming},
  journal={IEEE Transactions on Circuits and Systems for Video Technology},
  volume={32},
  number={11},
  pages={7505--7517},
  year={2022},
  publisher={IEEE}
}

@inproceedings{trufor,
  title={Trufor: Leveraging all-round clues for trustworthy image forgery detection and localization},
  author={Guillaro, Fabrizio and Cozzolino, Davide and Sud, Avneesh and Dufour, Nicholas and Verdoliva, Luisa},
  booktitle={Proceedings of the IEEE/CVF conference on computer vision and pattern recognition},
  pages={20606--20615},
  year={2023}
}

@article{mfi-net,
  title={MFI-Net: Multi-feature fusion identification networks for artificial intelligence manipulation},
  author={Ren, Ruyong and Hao, Qixian and Niu, Shaozhang and Xiong, Keyang and Zhang, Jiwei and Wang, Maosen},
  journal={IEEE Transactions on Circuits and Systems for Video Technology},
  volume={34},
  number={2},
  pages={1266--1280},
  year={2023},
  publisher={IEEE}
}

@inproceedings{sam,
  title={Segment anything},
  author={Kirillov, Alexander and Mintun, Eric and Ravi, Nikhila and Mao, Hanzi and Rolland, Chloe and Gustafson, Laura and Xiao, Tete and Whitehead, Spencer and Berg, Alexander C and Lo, Wan-Yen and others},
  booktitle={Proceedings of the IEEE/CVF international conference on computer vision},
  pages={4015--4026},
  year={2023}
}

@article{li2025beyond,
  title={Beyond Fully Supervised Pixel Annotations: Scribble-Driven Weakly-Supervised Framework for Image Manipulation Localization},
  author={Li, Songlin and Yu, Guofeng and Guo, Zhiqing and Diao, Yunfeng and Ma, Dan and Yang, Gaobo and Wang, Liejun},
  journal={arXiv preprint arXiv:2507.13018},
  year={2025}
}

@inproceedings{wscl,
  title={Towards generic image manipulation detection with weakly-supervised self-consistency learning},
  author={Zhai, Yuanhao and Luan, Tianyu and Doermann, David and Yuan, Junsong},
  booktitle={Proceedings of the IEEE/CVF International Conference on Computer Vision},
  pages={22390--22400},
  year={2023}
}

@article{EdgeCAM,
  title={Exploring weakly-supervised image manipulation localization with tampering edge-based class activation map},
  author={Zhou, Yang and Wang, Hongxia and Zeng, Qiang and Zhang, Rui and Meng, Sijiang},
  journal={Expert Systems with Applications},
  volume={249},
  pages={123501},
  year={2024},
  publisher={Elsevier}
}

@inproceedings{SOWCL,
  title={Self-Optimization Training for Weakly Supervised Image Manipulation Localization},
  author={Zhu, Zhangchen and Li, Jiafeng and Wen, Ying},
  booktitle={ICASSP 2025-2025 IEEE International Conference on Acoustics, Speech and Signal Processing (ICASSP)},
  pages={1--5},
  year={2025},
  organization={IEEE}
}

@article{WSCCL,
  title={Weakly-supervised cross-contrastive learning network for image manipulation detection and localization},
  author={Bai, Ruyi},
  journal={Knowledge-Based Systems},
  volume={310},
  pages={113033},
  year={2025},
  publisher={Elsevier}
}

@inproceedings{casia,
  title={Casia image tampering detection evaluation database},
  author={Dong, Jing and Wang, Wei and Tan, Tieniu},
  booktitle={2013 IEEE China summit and international conference on signal and information processing},
  pages={422--426},
  year={2013},
  organization={IEEE}
}

@inproceedings{nist16,
  title={MFC datasets: Large-scale benchmark datasets for media forensic challenge evaluation},
  author={Guan, Haiying and Kozak, Mark and Robertson, Eric and Lee, Yooyoung and Yates, Amy N and Delgado, Andrew and Zhou, Daniel and Kheyrkhah, Timothee and Smith, Jeff and Fiscus, Jonathan},
  booktitle={2019 IEEE Winter Applications of Computer Vision Workshops (WACVW)},
  pages={63--72},
  year={2019},
  organization={IEEE}
}

@article{columbia,
  title={Columbia uncompressed image splicing detection evaluation dataset},
  author={Hsu, J and Chang, SF},
  journal={Columbia DVMM Research Lab},
  volume={6},
  year={2006}
}

@inproceedings{coverage,
  title={COVERAGE—A novel database for copy-move forgery detection},
  author={Wen, Bihan and Zhu, Ye and Subramanian, Ramanathan and Ng, Tian-Tsong and Shen, Xuanjing and Winkler, Stefan},
  booktitle={2016 IEEE international conference on image processing (ICIP)},
  pages={161--165},
  year={2016},
  organization={IEEE}
}

@INPROCEEDINGS{imd2020,
author = {Novozamsky, Adam and Mahdian, Babak and Saic, Stanislav},
title = {IMD2020: A Large-Scale Annotated Dataset Tailored for Detecting Manipulated Images},
booktitle = {2020 IEEE Winter Applications of Computer Vision Workshops (WACVW)},
year = {2020},
month = {March},
pages = {71-80}
}

@article{cocoglide,
  title={Glide: Towards photorealistic image generation and editing with text-guided diffusion models},
  author={Nichol, Alex and Dhariwal, Prafulla and Ramesh, Aditya and Shyam, Pranav and Mishkin, Pamela and McGrew, Bob and Sutskever, Ilya and Chen, Mark},
  journal={arXiv preprint arXiv:2112.10741},
  year={2021}
}

@inproceedings{in-the-wild,
  title={Fighting fake news: Image splice detection via learned self-consistency},
  author={Huh, Minyoung and Liu, Andrew and Owens, Andrew and Efros, Alexei A},
  booktitle={Proceedings of the European conference on computer vision (ECCV)},
  pages={101--117},
  year={2018}
}

@article{dso,
  title={Exposing digital image forgeries by illumination color classification},
  author={De Carvalho, Tiago Jos{\'e} and Riess, Christian and Angelopoulou, Elli and Pedrini, Helio and de Rezende Rocha, Anderson},
  journal={IEEE Transactions on Information Forensics and Security},
  volume={8},
  number={7},
  pages={1182--1194},
  year={2013},
  publisher={IEEE}
}

@inproceedings{korus,
  title={Evaluation of random field models in multi-modal unsupervised tampering localization},
  author={Korus, Pawe{\l} and Huang, Jiwu},
  booktitle={2016 IEEE international workshop on information forensics and security (WIFS)},
  pages={1--6},
  year={2016},
  organization={IEEE}
}

@inproceedings{tinyvit,
  title={Tinyvit: Fast pretraining distillation for small vision transformers},
  author={Wu, Kan and Zhang, Jinnian and Peng, Houwen and Liu, Mengchen and Xiao, Bin and Fu, Jianlong and Yuan, Lu},
  booktitle={European conference on computer vision},
  pages={68--85},
  year={2022},
  organization={Springer}
}

@article{imdlbenco,
  title={Imdl-benco: A comprehensive benchmark and codebase for image manipulation detection \& localization},
  author={Ma, Xiaochen and Zhu, Xuekang and Su, Lei and Du, Bo and Jiang, Zhuohang and Tong, Bingkui and Lei, Zeyu and Yang, Xinyu and Pun, Chi-Man and Lv, Jiancheng and others},
  journal={Advances in Neural Information Processing Systems},
  volume={37},
  pages={134591--134613},
  year={2024}
}

@misc{forensichub,
      title={ForensicHub: A Unified Benchmark \& Codebase for All-Domain Fake Image Detection and Localization}, 
      author={Bo Du and Xuekang Zhu and Xiaochen Ma and Chenfan Qu and Kaiwen Feng and Zhe Yang and Chi-Man Pun and Jian Liu and Jizhe Zhou},
      year={2025},
      eprint={2505.11003},
      archivePrefix={arXiv},
      primaryClass={cs.CV},
      url={https://arxiv.org/abs/2505.11003}, 
}

@article{copy_move,
author = {He, Guangyang and Zhang, Xiang and Wang, Fan and Fu, Zhangjie},
title = {A novel copy-move detection and location technique based on tamper detection and similarity feature fusion},
year = {2024},
issue_date = {2024},
publisher = {Inderscience Publishers},
address = {Geneva 15, CHE},
volume = {17},
number = {6},
issn = {1754-8632},
url = {https://doi.org/10.1504/ijaacs.2024.142523},
doi = {10.1504/ijaacs.2024.142523},
journal = {Int. J. Auton. Adapt. Commun. Syst.},
month = jan,
pages = {514–529},
numpages = {15}
}

@article{deepfake1,
author = {Gu, Fei and Dai, Yunshu and Fei, Jianwei and Chen, Xianyi},
title = {Deepfake detection and localisation based on illumination inconsistency},
year = {2024},
issue_date = {2024},
publisher = {Inderscience Publishers},
address = {Geneva 15, CHE},
volume = {17},
number = {4},
issn = {1754-8632},
url = {https://doi.org/10.1504/ijaacs.2024.139383},
doi = {10.1504/ijaacs.2024.139383},
journal = {Int. J. Auton. Adapt. Commun. Syst.},
month = jan,
pages = {352–368},
numpages = {16}
}

\end{document}